%% file: main.tex
\documentclass[10pt,twocolumn,letterpaper]{article}

\usepackage{cvpr}              

\usepackage{float}
\usepackage{algorithm}
\usepackage{algorithmic}
\usepackage{amsmath,amssymb,eufrak,eucal}
\usepackage{latexsym}
\usepackage{framed,multirow}
\usepackage{color}
\usepackage{amsfonts}
\usepackage{bm}
\usepackage{latexsym}
\usepackage{url} 
\usepackage{times}
\usepackage{epsfig}
\usepackage{CJKutf8}

\usepackage{graphicx}
\usepackage{amsmath}
\usepackage{amssymb}
\usepackage{booktabs}

\usepackage{subcaption}

\usepackage[pagebackref,breaklinks,colorlinks]{hyperref}

\usepackage[capitalize]{cleveref}
\crefname{section}{Sec.}{Secs.}
\Crefname{section}{Section}{Sections}
\Crefname{table}{Table}{Tables}
\crefname{table}{Tab.}{Tabs.}

\newcommand\blfootnote[1]{%
  \begingroup
  \renewcommand\thefootnote{}\footnote{#1}%
  \addtocounter{footnote}{-1}%
  \endgroup
}

\def\cvprPaperID{5760} 
\def\confName{CVPR}
\def\confYear{2022}
\newcommand*{\affaddr}[1]{#1} 
\newcommand*{\affmark}[1][*]{\textsuperscript{#1}}
\newcommand*{\email}[1]{\texttt{#1}}

\begin{document}
\begin{CJK}{UTF8}{}
\title{Bi-level Alignment for Cross-Domain Crowd Counting}

\author{%
Shenjian Gong\affmark[1], Shanshan Zhang\affmark[1,*], Jian Yang\affmark[1], Dengxin Dai\affmark[2], and Bernt Schiele\affmark[2]\\
\affaddr{\affmark[1]PCA Lab, Key Lab of Intelligent Perception and Systems for High-Dimensional Information \\of Ministry of Education, and Jiangsu Key Lab of Image and Video Understanding for Social Security,\\ School of Computer Science and Engineering, Nanjing University of Science and Technology
}\\
\affaddr{\affmark[2]MPI Informatics}\\
\email{\{shenjiangong,shanshan.zhang,csjyang\}@njust.edu.cn}\\ \email{\{ddai,schiele\}@mpi-inf.mpg.de}
}

\maketitle
\blfootnote{*Corresponding author}
\begin{abstract}
   Recently, crowd density estimation has received increasing attention. The main challenge for this task is to achieve high-quality manual annotations on a large amount of training data. To avoid reliance on such annotations, previous works apply unsupervised domain adaptation (UDA) techniques by transferring knowledge learned from easily accessible synthetic data to real-world datasets.
   However, current state-of-the-art methods either rely on external data for training an auxiliary task or apply an expensive coarse-to-fine estimation.
   In this work, we aim to develop a new adversarial learning based method, which is simple and efficient to apply.
   To reduce the domain gap between the synthetic and real data, we design a bi-level alignment framework (BLA) consisting of (1) task-driven data alignment and (2) fine-grained feature alignment. In contrast to previous domain augmentation methods, we introduce AutoML to search for an optimal transform on source, which well serves for the downstream task. On the other hand, we do fine-grained alignment for foreground and background separately to alleviate the alignment difficulty. We evaluate our approach on five real-world crowd counting benchmarks, where we outperform existing approaches by a large margin. Also, our approach is simple, easy to implement and efficient to apply. The code is publicly available at \url{https://github.com/Yankeegsj/BLA}.
\end{abstract}

\input{illustration/da_dr_dl}

\section{Introduction}
Crowd counting aims to estimate the number of persons in crowded scenes. This task has gained a lot of attention \cite{NAS_count,gspt} 
as it is useful for video surveillance, traffic control and human behavior analysis. 
Especially under pandemics such as COVID-19, it can be used to monitor and regulate the flow of people for safety reasons.

In recent years, many methods have been proposed for crowd counting. Most state-of-the-art approaches rely on ground truth density maps for a large number of training images, with each human head marked. However, it is extremely expensive to annotate so many human heads and for high-density regions such manual labels can be noisy. To reduce annotations costs, a large-scale synthetic dataset GTA5 Crowd Counting (GCC) \cite{GCC} was created, serving as a well-established training set with automatic annotations. However, models trained on the synthetic dataset perform poorly due to the large domain gap between synthetic and real-world images. Thus, it is necessary to investigate how to adapt the models trained on the synthetic domain to the real domain, without requiring annotations on the latter, i.e. via unsupervised domain adaptation (UDA).


There are a few UDA methods proposed for crowd counting. For instance, SE Cycle-GAN \cite{GCC} translates synthetic images to the real domain with improved Cycle-GAN and then trains purely on the translated images; Gaussian Process-based iterative learning (GP) \cite{GP} generates pseudo-labels on the target images via a Gaussian process to allow for supervised training on the target domain. More recently,  
better performance has been achieved by employing an adversarial framework to align features from both source and target domains \cite{FSC, FADA}. However, FSC \cite{FSC} introduces an auxiliary task of semantic segmentation, relying on external labeled human body segmentation datasets for pre-training;
FADA \cite{FADA} performs a coarse-to-fine estimation, making the inference less efficient.

In this paper, we aim to develop a new adversarial learning based method, which is more effective and flexible. We investigate the key components to boost performance. Previous methods employed either domain randomization or style transfer for source domain augmentation, using no priors or the target style priors only. In contrast, our task-driven data alignment is able to control the domain augmentation based on both the target style priors and the task objective such that it is optimized for our crowd counting task on the given target domain.
We show a comparison of three different source domain augmentation methods in Fig. \ref{DA_DR_dlv}.
On the other hand, since the foreground and background regions differ significantly in semantics, we propose a fine-grained feature alignment to handle them separately.



To summarize, the contributions of our work are as follows: (1) For more effective and efficient synthetic to real adaptive crowd counting, we propose a novel adversarial learning based method, consisting of bi-level alignments: task-driven data alignment and fine-grained feature alignment. (2) To the best of our knowledge, it is the first UDA approach to search for the optimal source data transform based on the downstream task performance on the target domain.
(3) Experimental results on various real datasets show that our method achieves state-of-the-art results for synthetic-to-real domain adaptation; also, our method is simple and efficient to apply.

\section{Related Works}
Since we solve the problem of domain adaptive crowd counting, we first review recent works; our major contribution is a novel domain augmentation method via AutoML, so we also discuss related works in the above two areas.
\subsection{Domain Adaptive Crowd Counting}


There are two groups of domain adaptation works in crowd counting: real-to-real and synthetic-to-real. Real-to-real adaptation aims to generalize models across real scenarios \cite{he2021error,CODA}, but since one real-world dataset is taken as the source domain, manual annotations are still needed. In contrast, synthetic-to-real adaptation fully avoids the requirement for manual annotations and thus is more interesting. In this work, we focus on synthetic-to-real adaptation.
One direct way is to translate the labeled synthetic images to the style of the real images and then train on the translated images \cite{GCC,gao2019domain}, but it is limited by the performance of the translation method. Another intuitive way is to generate pseudo-labels on the target real images to allow for supervised learning on the target domain \cite{GP,gao2019domain}. More recently, the adversarial framework has been leveraged to achieve better performance via feature alignment between source and target domains \cite{FSC, FADA}. However, previous works are not efficient, requiring external training data or additional inference time. For instance, FSC \cite{FSC} introduces an auxiliary task of semantic segmentation, relying on external labeled human body segmentation datasets for pre-training; FADA \cite{FADA} performs a coarse-to-fine estimation, making the inference less efficient.
In this work, we leverage adversarial training but aim to develop a simple yet effective method.

\subsection{Domain Augmentation}
Previously, there are two ways to augment the existing source domain: one is domain randomization, randomly changing the style of source images; the other is style transfer, translating the source images to the target style. Both of them preserve the contents of source images to allow for supervised learning.

Domain randomization is mostly used for domain generalization, which handles unknown target domains. The generated new samples with random styles are helpful to enhance the generalization ability of the trained model.
One direct way is to generate various styles of images by style transfer CNN \cite{DR_style} or data augmentation \cite{DR_IROS}. 
On the other hand, AdaIN \cite{AdaIN} has demonstrated that the mean and variance of convolutional feature maps can be used to represent the image style, making the domain randomization easier \cite{reducing}.
Some recent works modify styles in the frequency domain \cite{AFourierbasedFramework,FSDR} by randomizing the domain-variant components, which represent styles. 

In contrast, style transfer is used for domain adaptation, where the target domain images are given to provide style guidance.
The most common way is to generate source-to-target images through Cycle-GAN \cite{cycada,GCC,gao2019domain}. 
Similar to domain randomization, some works also do it in the frequency domain by replacing the style-related components with target ones, e.g. the amplitude of Fourier transform \cite{FDA}.



In this work, we propose a novel domain augmentation method named task-driven data alignment which is superior than domain randomization and style transfer. The major difference is that the augmentation is controlled by both the target style priors and by verifying the counting performance on a target-like domain.
In this way, we are able to optimize augmentation for our crowd counting task on the given target domain.





\input{illustration/method}




\subsection{Auto Machine Learning}
Auto machine learning (AutoML) aims to free human practitioners and researchers from selecting the optimal values for each hyperparameter, such as learning rate, weight decay, and dropout \cite{AutoHAS}, or designing well-performing network architectures \cite{chen2021one}. 
Pioneers in this field develop optimization methods to guide the search process based on reinforcement learning (RL) \cite{autogan}, evolutionary algorithm (EA) \cite{lighttrack} and Bayesian optimization \cite{liu2018progressive}. These works are often impractical because of the required computational overhead. In contrast, a differentiable controller \cite{NAO} converts the selection into a continuous hidden space optimization problem, allowing for an efficient search process performed by a gradient-based optimizer.

We apply a differentiable controller to search for several hyperparameters which represent styles. The transformed source images are then used for training, where the feature alignment becomes easier as the domain gap is reduced. 
In order to verify the searched hyperparameters, we construct a target-like domain in the feature space via AdaIN.

    \section{Bi-Level Alignment Counting (BLA)}

In this section, we will introduce our bi-level alignment method for cross-domain crowd counting. Our core idea is to perform alignment between the source and target domains at both data-level and feature-level via two components namely task-driven data alignment and fine-grained feature alignment.
The overall pipeline is depicted in Fig. \ref{method} and detailed descriptions are provided in the following.

\subsection{Problem Formulation}

For UDA crowd counting, we have an annotated synthetic dataset $S=(x_S^i,y_S^i)^{N_{S}}_{i=1}$ as source and an unlabeled real-world dataset $T={(x_T^i)}^{N_{T}}_{i=1}$ as target, where $x_S^i,x_T^i \in \mathbb{R}^{3 \times H \times W}$ denote an arbitrary image from the source and target domain, and $y_S^i \in \mathbb{R}^{H \times W}$ represents the ground truth density map in source. 
Our goal is to obtain a model that performs well on the target domain via reducing the large domain gap between the source and target. 

\subsection{Overview}
As shown in Fig. \ref{method}, we propose a new UDA method for crowd counting based on adversarial learning. It consists of feature extractor $(\mathcal{F})$, density estimator $(\mathcal{E})$, task-driven data alignment and local fine-grained discriminator $(\mathcal{D})$.

At training time, the source dataset $S$ is first transformed to $S^+$, with the same labels;
a pair of images $(x_{S^+},x_T)$ from the augmented source and target domains are fed into $\mathcal{F}$, obtaining corresponding feature maps $(F_S,F_T)$, $F_S$ and $F_T\in \mathbb{R}^{C\times h \times w}$;
$\mathcal{D}$ performs feature alignment by passing reversed gradients to $\mathcal{F}$;
in the end, $\mathcal{E}$ predicts density maps $\widetilde y_S$ based on $F_S$, supervised by $y_{S}$. At test time, the inference is rather simple: each target image $x_T^i$ only goes through $\mathcal{F}$ and $\mathcal{E}$ to obtain the predicted density map $\widetilde y_T$.

Following previous works, we employ VGG16 \cite{VGG} as our feature extractor $\mathcal{F}$. For $\mathcal{E}$, we stack a series of convolution and deconvolution layers, inspired by \cite{gao2019domain}.

The density estimation loss $\boldsymbol{\mathcal{L}}_{E}$ on a labeled source image can be defined as follows: 



\begin{equation}
\boldsymbol{\mathcal{L}}_{E}= \sum \left \| y_S-\widetilde y_S  \right \| ^2.
\end{equation}
\subsection{Task-Driven Data Alignment}


Typically, domain augmentation enhances the performance on the target domain using generated new samples via domain randomization or style transfer. In contrast to previous domain augmentation methods that are blind to the downstream task, our method searches for the most suitable augmentation based on both the target styles and the task performance on target via AutoML. In such a task-driven way, our method is expected to find a transform that better serves for the downstream task on the given target domain.

It has been shown in \cite{gspt} that for the crowd counting task, images mainly differ in color, scale and perspective. Accordingly, in this paper, we define three basic transform units: RGB2Gray, scaling and perspective transform.
Each transform
is a combination of the above three transform units.
As shown in Tab.~\ref{table:search_space}, one transform is defined by five parameters, among which only three are searched for simplicity. 
It is notable the transform is not limited to the above three units and can be easily extended to different and also more types, with proper manual definitions. 

\input{illustration/search_space}

Given a transform, we split the whole source set into several subsets via a transform tree, as shown in Fig. \ref{split}. 
At the 1st level, the whole source dataset is split into two subsets with a ratio of p$_\text{G}$, i.e. some images are converted to gray scale images (along path \textbf{Y}), while others are kept the same (along path \textbf{N}).
At the following levels, each subset generated from the previous level goes through one split with a given split ratio and an attribute parameter. Given three transform units, we finally generate 8 ($2^3$) subsets.


\input{illustration/split}
In this paper, we use a differential controller to guide the direction of our search process. The search process is iterated via multiple rounds, each of which is described in Alg. \ref{data-level alignment}.
At each round, we first transform the source data given some transform candidates, and then obtain the reward of each transform via validation on a generated target-like set; after that, we learn the mapping function from transforms to corresponding rewards via training a differentiable controller; finally, we update the transform candidates based on the controller and goes to the next search round.
In the following, we explain the above process in more detail.

\textbf{Source Data Transformation Based on Candidate Transforms.} At this stage, we first randomly initialize a transform set $\mathbb{D}={(d_k)}^{N_{\mathbb{D}}}_{k=1}$.
Given an arbitrary transform $d_k$, we split the whole source set into several subsets via a transform tree, as illustrated in Fig. \ref{split}. 
In this way, we apply one transform  on the source data and obtain a new mixed source dataset $S^+_k=(x_{S^+_k}^i,y_{S^+_k}^i)^{N_{S}}_{i=1}$.
Due to the large number of source images, we do not do standard data augmentation, but apply the transform on the original data to keep the same size of $S^+_k$ and $S$.


\textbf{Candidate Transform Validation.} 
Based on each new source dataset $S^+_k$, we train the whole network as shown in Fig. \ref{method} with the learning objective from Eq. \ref{objective}.
Now we need to evaluate the alignment quality of each $S^+_k$. Ideally, this should be done by measuring the counting performance on $T$.
Unfortunately, we do not have labels for $T$. To address this problem, we propose a validation feature generator, which takes the features of a pair of source and target image features $(F_{S},F_T)$ from the feature extractor as input and generate a new feature $F_V$ via AdaIN \cite{AdaIN}, which is a mixture of source contents and target style, namely a target-like image feature.

Specifically, we first compute the source style representation with channel-wise mean and standard deviation $\mu (F_{S})$, $\sigma (F_{S}) \in \mathbb{R}^C$ and the target style representation $\mu (F_{T})$, $\sigma (F_{T}) \in \mathbb{R}^C$.
Then we replace the style of $F_{S}$ with that of $F_{T}$ and obtain $F_V$:
\begin{equation}
F^c_{V}=\mu (F^c_{T})+\sigma (F^c_{T})\cdot (\frac{F^c_{S}-\mu (F^c_{S})}{\sigma (F^c_{S})}),
\end{equation}
where $c\in \{1,2,3, \cdot \cdot \cdot C\}$ is the channel index. After that, we feed $F_{V}$ to the density estimator and get $\widetilde y_{V}$. Because $F_{S}$ and $F_{V}$ share the same contents, we evaluate $\widetilde y_{V}$ based on $y_S$. 
In this way, we obtain the evaluated validation performance $p_{k}$ as reward for transform $d_k$.


\textbf{Candidate Transform Update.} After obtaining the reward for each transform in $\mathbb{D}$, we then train a differentiable controller and let it learn the mapping function from a transform to its corresponding reward. The controller is of encoder-decoder structure. The encoder takes a transform as input, maps it to a hidden state, 
and predicts its performance as $\widetilde p_{k}$. The decoder reconstructs the transform $d_{k}$ as $\widetilde d_{k}$ from the hidden state. 
The loss function of our controller is defined as:
\begin{equation}
\boldsymbol{\mathcal{L}}_{C}=\left \| d_{k}-\widetilde d_{k} \right \| ^2 + \left \| p_{k}-\widetilde p_{k} \right \| ^2.
\label{equation:controller}
\end{equation}
Same with NAO \cite{NAO}, we then update the hidden state towards the gradient direction of improved performance and obtain a new transform set $\mathbb{D'}$, for better alignment.
After several rounds, we choose the optimal transform from all validated transforms based on their rewards. Please refer to \cite{NAO} for more details regarding the update procedure.

\input{illustration/search_process}
\subsection{Fine-Grained Feature Alignment} 
To perform feature alignment, we employ adversarial learning via a discriminator and a gradient reverse layer.
Inspired by the success of using segmentation as an auxiliary task for crowd counting 
\cite{inverse}, 
we propose a fine-grained discriminator, with two separated classification heads for foreground and background regions.
To handle the unbalanced numbers of foreground and background pixels, the discriminator is applied on local patches instead of pixels.


Given the grid size $G=(g_h,g_w)$, we feed a pair of feature maps $(F_S,F_T)$ to $\mathcal{D}$ and obtain two pairs of patch-level discrimination maps: $(O_{FS},O_{BS})$ for source and $(O_{FT},O_{BT})$ for target, separating foreground and background. Each map $O_{FS},O_{FT},O_{BS},O_{BT} \in \mathbb{R}^{(H/g_{h})\times (W/g_{w})}$.

The segmentation masks ($M_S, \widetilde M_T$) are obtained by thresholding the ground truth density maps. Please note that for the target we use pseudo density maps instead.
Specifically, we apply a threshold of $th$ on each patch, to threshold the sum of all its pixel values.




As shown in Fig. \ref{method}, we define local fine-grained discrimination losses of background and foreground
$\boldsymbol{\mathcal{L}}_{B}$, $\boldsymbol{\mathcal{L}}_{F}$, as follows:


\begin{equation}
\begin{split}
\boldsymbol{\mathcal{L}}_{D}&=\boldsymbol{\mathcal{L}}_{B}+\boldsymbol{\mathcal{L}}_{F}\\
\boldsymbol{\mathcal{L}}_{B}&=\boldsymbol{\mathcal{L}}_{BS}+\boldsymbol{\mathcal{L}}_{BT}\\
&=\sum -(1-M_S) \cdot log(1-O_{BS}) \\
&+\sum -(1-\widetilde M_{T}) \cdot log(O_{BT}),\\
\boldsymbol{\mathcal{L}}_{F}&=\boldsymbol{\mathcal{L}}_{FS}+\boldsymbol{\mathcal{L}}_{FT}\\
&=\sum -M_S\cdot log(1-O_{FS})\\
&+\sum -\widetilde M_{T} \cdot log(O_{FT}).
\end{split}
\label{LFD_loss}
\end{equation}
%
%
We use the same back-propagation optimizing scheme with the gradient reverse layer \cite{grl} for adversarial learning. 


\subsection{Optimization}
The optimization objective of the whole method is:


%
\begin{equation}
\boldsymbol{\mathcal{L}}=\boldsymbol{\mathcal{L}}_{E}+ \lambda\boldsymbol{\mathcal{L}}_{D},
\label{objective}
\end{equation}
where $\lambda$ is a weight factor to balance the task loss $\boldsymbol{\mathcal{L}}_{E}$ and the domain adaptation loss $\boldsymbol{\mathcal{L}}_{D}$.


The whole network is optimized via two steps.
For data alignment, we first optimize all the parameters in the network including the feature alignment component for each transform to obtain the corresponding reward,
during the search process in Alg. \ref{data-level alignment}. 
After selecting the best transform, we retrain the whole network including the feature alignment component with the transformed $S^+$.

\section{Experiments}
We first introduce the datasets, evaluation metrics and implementation details; then we provide comparisons with state-of-the-art methods, followed by analysis on data alignment; finally, we perform some ablation studies.

\subsection{Datasets and Evaluation Metrics}
To evaluate the proposed method, the experiments are conducted under adaptation scenarios from GCC \cite{GCC} to five large-scale real-world datasets, i.e. ShanghaiTech Part A/B (SHA/SHB) \cite{SHA}, QNRF \cite{QNRF}, UCF-CC-50 \cite{UCF50} and WorldExpo’10 \cite{WE} respectively. Statistics are listed in Tab. \ref{table:datasets}.

\input{illustration/datasets}

Following previous works, we adopt Mean Absolute Error (MAE) and Mean Squared Error (MSE) as evaluation metrics.
They are formulated as: $MAE=\frac{1}{N} \sum_{i=1}^{N} \left | \sum y_i-\sum\widetilde{y}_i  \right |,$ and $MSE= \sqrt{\frac{1}{N} \sum_{i=1}^{N} {\left | \sum y_i-\sum\widetilde{y}_i  \right |}^2},$
where $N$ is the number of test images; $\sum y_i$, $\sum\widetilde{y}_i$ represent the ground truth and predicted number on the $i$-th image respectively.

\input{illustration/STOA}
\subsection{Implementation Details}


The architectures of feature extractor ($\mathcal{F}$), density estimator ($\mathcal{E}$), fine-grained discriminator ($\mathcal{D}$) and controller ($\mathcal{C}$) are listed in supplementary. 
We input 4 pairs of source and target images with a uniform size of $576 \times 768$ at each iteration. Following the previous work \cite{gao2019domain}, we generate the ground truth density map using Gaussian kernel with a kernel size of 15$\times$ 15 and a fixed standard deviation of 4. 
We set $th$, $G$ in Eq. \ref{LFD_loss} and $\lambda$ in Eq. \ref{objective} to 0.005, (16,16) and 1.0 respectively for all our experiments; both p$_\text{S}$ and p$_\text{PT}$ are set to 0.5. And the gradient factor of the gradient reverse layer is set to 0.01. 
We also adopt a scene regularization strategy proposed by \cite{GCC} to avoid negative knowledge transfer. We train the adversarial framework and controller with Adam optimizer with default parameters, and their learning rates are initialized as $10^{-5}$ and $10^{-1}$ respectively. All experiments are conducted on a single NVIDIA RTX 2080TI GPU with 11GB of VRAM and our code is implemented with Pytorch.


\subsection{Comparisons with State-of-the-Art}
We compare our method BLA with previous published unsupervised domain adaptive crowd counting methods under the adaptation scenarios from synthetic GCC dataset to five different real-world datasets. All methods employ VGG16 \cite{VGG} as backbone.

\input{illustration/vision_results}
From the results in Tab. \ref{table:STOA}, we have the following observations:
(1) Our proposed method outperforms all existing domain adaptation methods by a large margin across different datasets and on WorldExpo'10 we achieve comparable results with DACC. In particular, on SHA our proposed method achieves 99.3 MAE and 145.0 MSE, outperforming previous best results by 13.1 pp w.r.t. MAE and 31.9 pp w.r.t. MSE. (2) Our method is robust across various real target datasets, showing high adaptability. As shown in Tab. \ref{table:datasets}, although the density of these real-world datasets varies a lot, we perform the best on all target domains.

\subsection{Analysis on Task-Driven Data Alignment}

In order to understand how the data alignment behaves, we show the searched transforms for different target datasets in Tab. \ref{table:searched_d}. It shows that our task-driven data alignment is quite interpretable, representing various domain gaps between source and different target domains.
For instance, GCC contains highly-saturated color images while UCF-CC-50 and WorldExpo'10 contain lots of images with low saturations, so the RGB2Gray ratios on them are rather high (0.85 and 0.98) as gray scale images are of 0 saturation; in contrast, SHB is closer to GCC in terms of saturation, so the RGB2Gray ratio on SHB is rather low (0.16). Similarly, since UCF-CC-50 is denser than other datasets, its scaling factor is particularly smaller, such that denser regions with small-scale heads will be generated.

\input{illustration/searched_d}
\input{illustration/different_dataset}

Moreover, we perform cross dataset validation by applying the searched transform on one dataset to another. As shown in Tab. \ref{table:different_dataset}, when testing on SHB, the transform searched on SHA underperforms that searched on SHB; and vice versa. These results indicate that each dataset requires its personalized transform to achieve the optimal performance. It is necessary to search for the most suitable transform on each dataset in an automatic way so as to avoid tedious manual designs.

Additionally, Fig. \ref{vision_results} shows some qualitative results on the SHA dataset. From Column 3, we can see that without adaptation, the model either fails to detect the presence of people in some areas (top row), or fails to get a correct estimate of the local density (middle and bottom rows). 
Our task-driven data alignment helps to reduce the errors largely, indicating the domain gaps are significantly narrowed. 
After further adding our fine-grained feature alignment, our BLA method provides more accurate final counts.


\subsection{Ablation Studies}
In this subsection, we conduct some ablation studies to analyze different components of our proposed BLA. All experiments are conducted under the GCC $\xrightarrow{}$ SHA adaptation due to its large variation in crowd density.

\textbf{Effects of Two Levels of Alignment.}
We first analyze the effects of data alignment and feature alignment. As shown in Tab. \ref{table:ablation}, the performance is significantly improved by $\sim$25 pp w.r.t MAE (from 134.7 to 109.1) when task-driven data alignment is employed. 
On the other hand, we also observe a large improvement of $\sim$14 pp w.r.t MAE (from 134.7 to 121.1) by replacing global feature alignment with fine-grained feature alignment.
Moreover, we obtain a total gain of $\sim$35 pp w.r.t MAE by adding both alignments. 
These results indicate the effects of two levels of alignment and the complementarity between them.

\input{illustration/ablation}

\textbf{Effect of Task-Driven Data Alignment}
To evaluate the effectiveness of task-driven data alignment, we replace it with domain randomization and style transfer. From Tab. \ref{table:drstdla}, we can see that our task-driven data alignment outperforms previous two data augmentation methods by a large margin.

\input{illustration/drstdla}



\textbf{Effect of Validation Feature Generator.}
To evaluate the alignment quality of each transform, we generate a target-like feature set for validation via AdaIN.
In Fig. \ref{adain}, we compare the counting performance on our generated validation set and the real target training set w.r.t. MAE, where the index indicates different combinations of transform parameters. We can see that the two curves go in a similar trend, i.e. the worst performance happens at index 0, the best performance happens at index 7, and there is fluctuation in between. This comparison demonstrates that our generated validation set is of high similarity to the real target set, allowing us to do effective validation without relying on target annotations.
On the other hand, we observe the performance varies a lot along the choice of transform parameters, showing that different transforms highly affect the performance. Thus it is of great importance to search for an optimal transform for a given target set.
\input{illustration/adain}
\textbf{Impact of Grid Size in Fine-Grained Feature Alignment.}
Our fine-grained feature alignment strategy is conducted in a patch-wise style. We analyze how the grid size $G$ affects the performance. As shown in Tab. \ref{table:grid}, if the size $G$ is too small or too large, there will be a data imbalance between the numbers of background and foreground patches, which results in poor feature alignment. Since the patch size of 16 performs the best, we use 16 as the default size in all experiments.

\input{illustration/grid}

In the supplementary material, we provide more ablation studies on the impact of segmentation threshold, effect of additional style transfer from $S^+$ to $T$, effect of using more transformations, and a comparison to grid search.


\section{Conclusion}
In this work, we propose a bi-level alignment framework for synthetic-to-real UDA crowd counting.
On one hand, we propose task-driven data alignment to search for a specific transform given the target set, which is applied on the source data to narrow down the domain gap at the data level.
On the other hand, to alleviate the alignment difficulty on the entire image, we propose to perform fine-grained feature alignment on foreground and background patches separately.
Extensive experiments on five real-world crowd counting benchmarks have demonstrated the effectiveness of our contributions.
\section*{Acknowledgements}
This work was supported in part by the  “111” Program B13022, Fundamental Research Funds for the Central Universities (No. 30920032201) and the National Natural Science Foundation of China (Grant No. 62172225).

\newpage

{\small
\bibliographystyle{ieee_fullname}
\bibliography{egbib}
}

\newpage

\end{CJK}
\end{document}

%% file: illustration/da_dr_dl.tex
\begin{figure} \centering
    \begin{subfigure}[a]{\linewidth}
    
    \begin{minipage}[t]{1\textwidth}
    \centering
\includegraphics[width=0.75\textwidth]{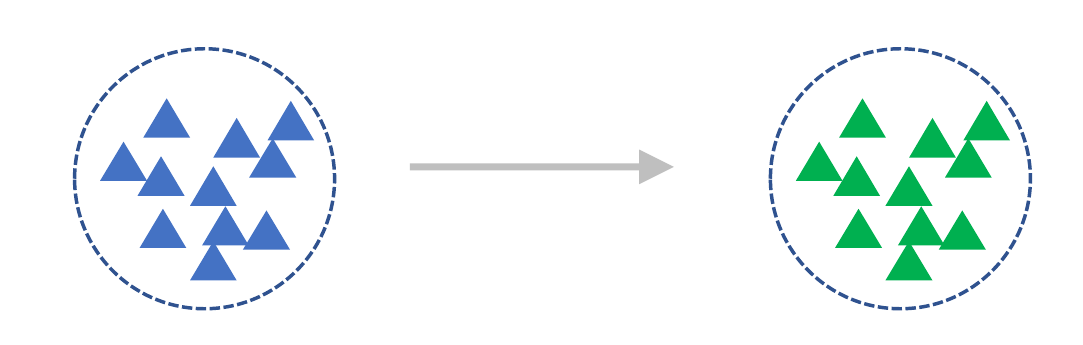}
\caption{Style Transfer}
\end{minipage}
\begin{minipage}[t]{1\textwidth}
\centering
\includegraphics[width=0.9\textwidth]{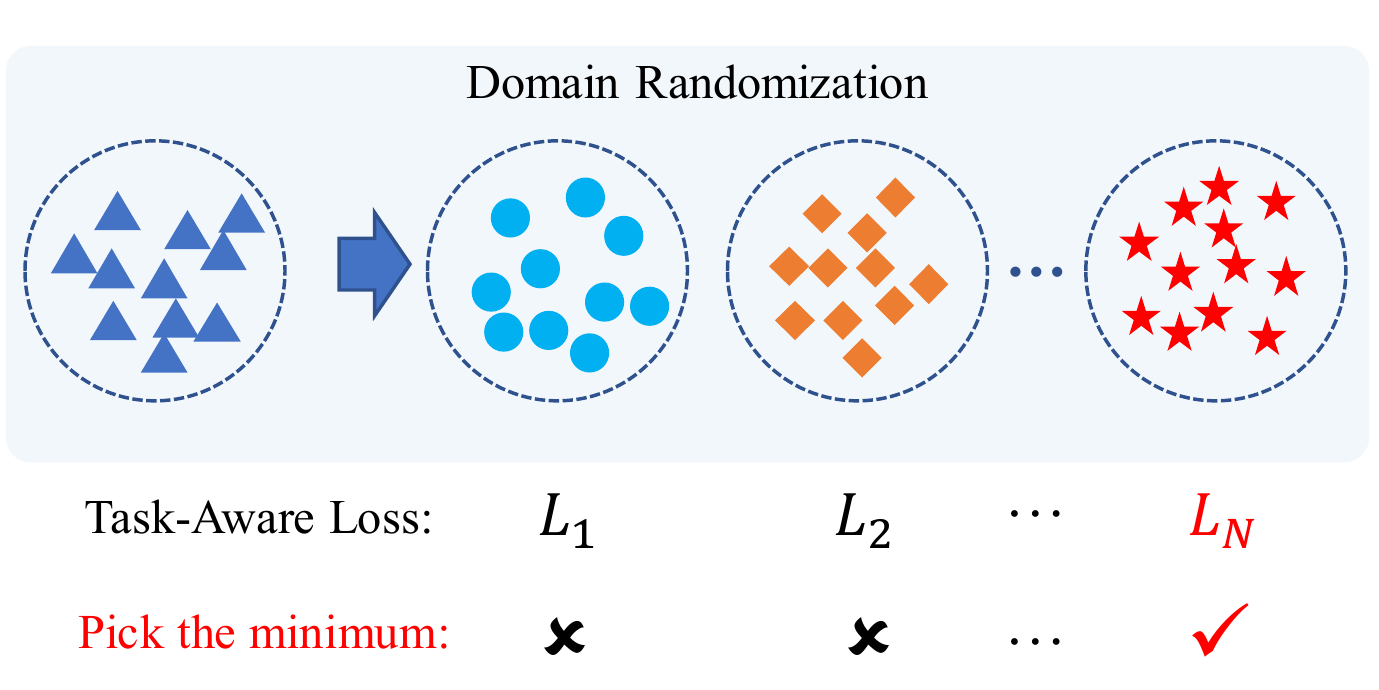}
\caption{Domain Randomization and Task-Driven Data Alignment}
\end{minipage}
    \end{subfigure} %
\caption{Comparison of three different ways for source domain augmentation. (a). Style transfer translates images from source to a target-like domain based on target style priors, but the translation is usually limited to color changes, and blind to the task objective. (b). Domain randomization augments the source domain randomly in a more diverse manner (colors, scales, etc.) but without any priors from target; our proposed task-driven data alignment is more similar to domain randomization; but instead of random selection, we pick the most suitable augmentation based on the task objective, which enables a more dynamic and robust model to the target domain.}
\label{DA_DR_dlv}
\end{figure}

%% file: illustration/method.tex
\begin{figure*}
	\begin{center}
		\includegraphics[width=0.91\textwidth]{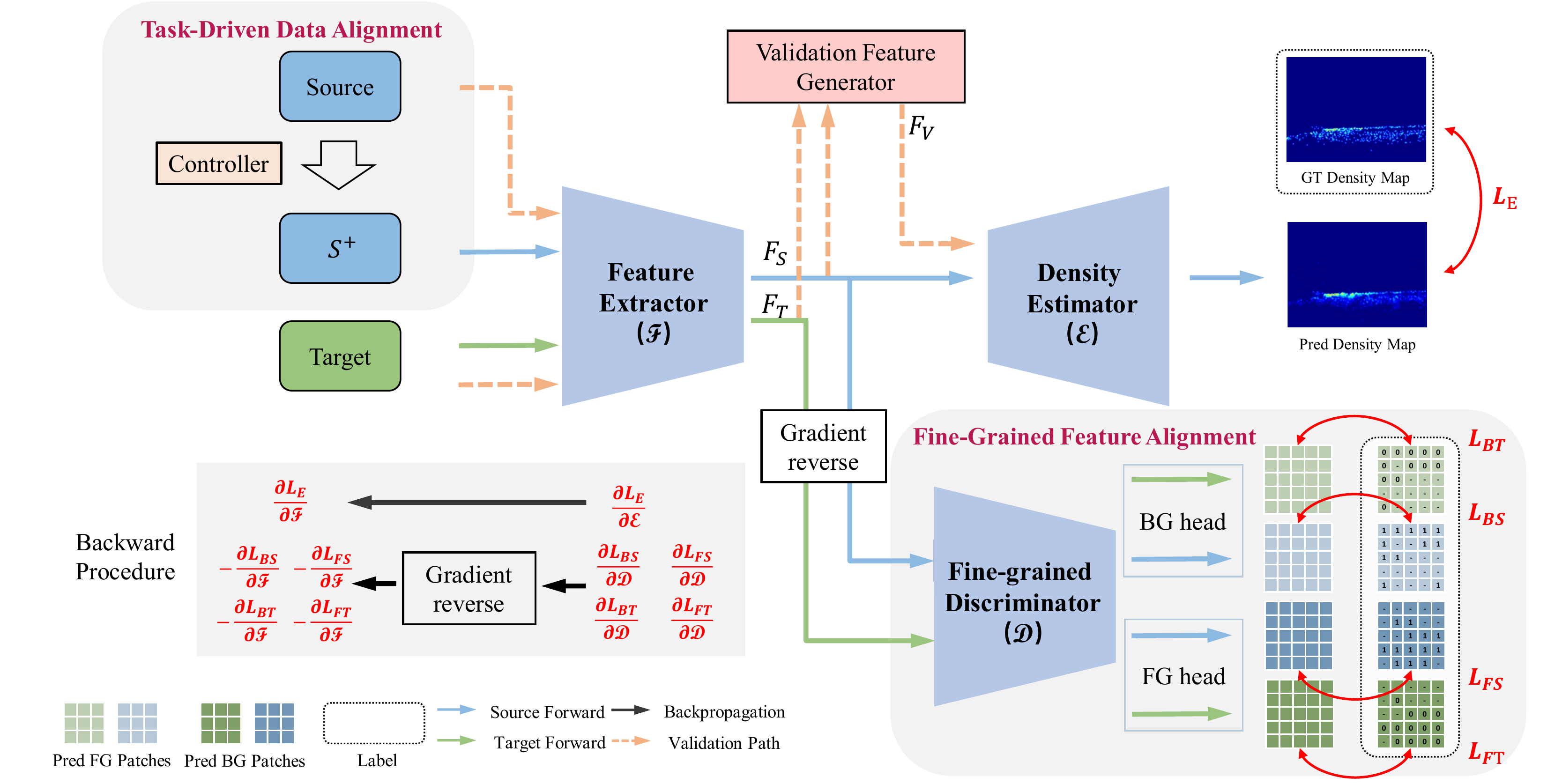}
	\end{center}
	\caption{Overview of our proposed bi-level alignment framework (BLA), which mainly consists of four components: feature extractor $(\mathcal{F})$, density estimator $(\mathcal{E})$, task-driven data alignment and local fine-grained discriminator $(\mathcal{D})$. At training time, the source dataset $S$ is transformed to $S^+$ with the optimal transform searched via task-driven data alignment (Alg. \ref{data-level alignment}), during which the validation feature generator provides target-like features for candidate transform validation.
	Then the entire network is optimized based on $S^+$ and $T$ using the training objective in Eq. \ref{objective}.
	\textbf{At test time, we simply feed a target image $x_T^i$ to $\mathcal{F}$ and $\mathcal{E}$ to obtain the predicted density map $\widetilde y_T$}. 
	}
	\label{method}
\end{figure*}

%% file: illustration/search_space.tex

\begin{table}[h]
\small
\renewcommand*{\arraystretch}{1.2}
	\centering
	\begin{tabular}{ccc}
		\toprule
		\multirow{2}{*}{Transform unit}  &\multicolumn{2}{c}{Parameters}\\ \cline{2-3} 
		 ~ &Split ratio &  Attribute\\
		\toprule
		RGB2Gray & p$_\text{G}$* &- \\
		Scaling & p$_\text{S}$& Scale factor*  \\
		Perspective transform & p$_\text{PT}$& Angle* \\
		\toprule
	\end{tabular}
	\caption{Each transform consists of three different units, each represented by two parameters: one for split ratio and another for attribute. * marks those parameters we search while others are fixed. A full transform set is generated by iterating each parameter.
	}
	\label{table:search_space}
\end{table}

%% file: illustration/split.tex

\begin{figure}
	\begin{center}
		\includegraphics[width=0.46\textwidth]{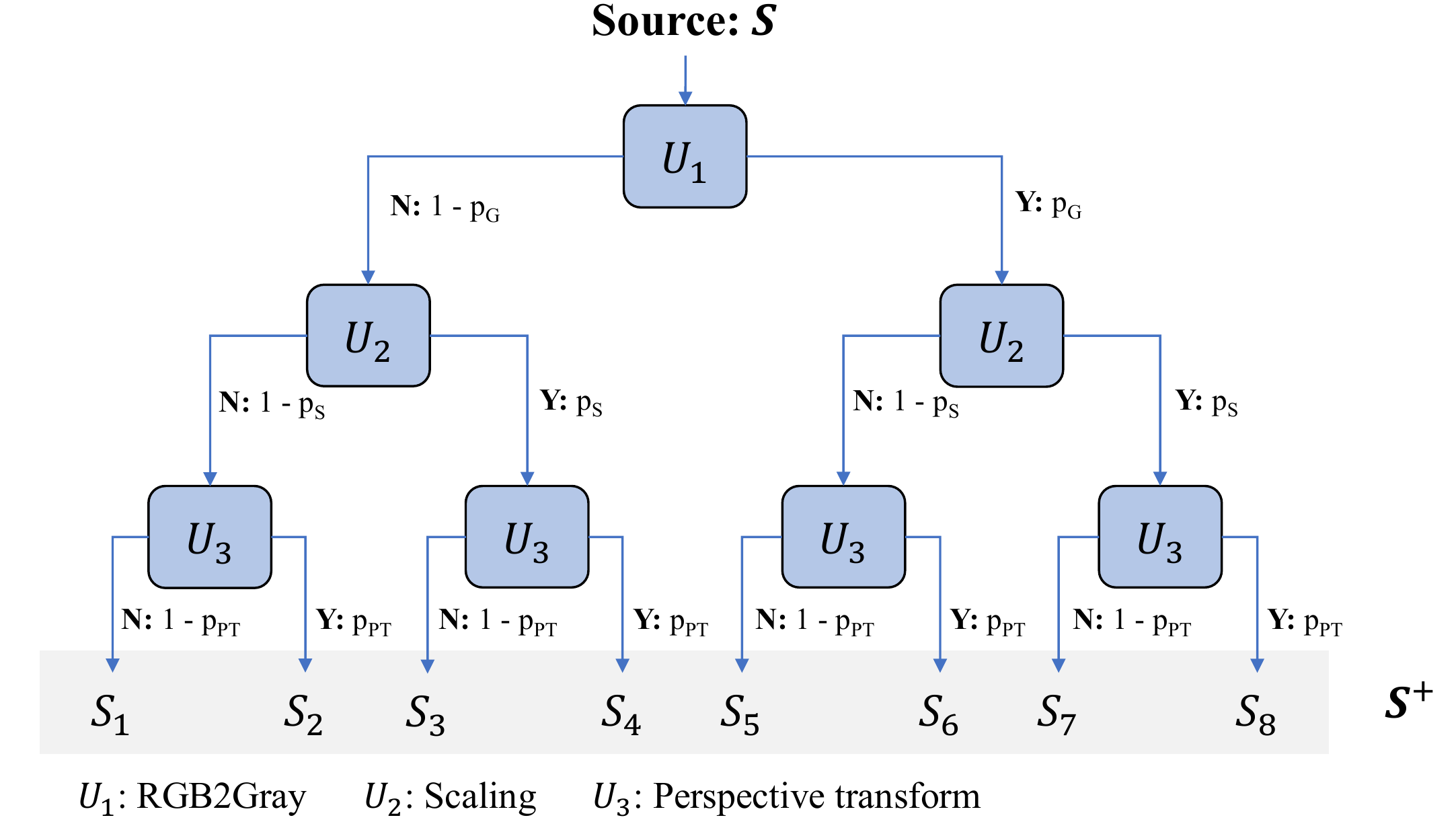}
	\end{center}
	\caption{We split the whole source set into several subsets via a transform tree according to p$_\text{G}$, p$_\text{S}$ and p$_\text{PT}$.
 ``\textbf{Y:} p$_\text{G}$'' refers to performing the transformation G with a ratio of p$_\text{G}$; ``\textbf{N:} 1 - p$_\text{G}$'' means that no transformation is performed with a ratio of 1 - p$_\text{G}$.
	Finally, we obtain the corresponding subsets as shown at the bottom row, which constitute $S^+$.
	}
	\label{split}
\end{figure}

%% file: illustration/search_process.tex
\begin{algorithm}[!tb]  
	\caption{Pseudo code of one-round search procedure of data-level alignment} 
	\label{data-level alignment}
	\textbf{Input}: Source and target domain training set $S$, $T$; the pre-trained  source only network $\mathcal{\widehat{G}}=\{\widehat{\theta}_{\mathcal{F} },\widehat{\theta}_{\mathcal{E}}\}$; the candidate transform set  $\mathbb{D}$ and the controller $\mathcal{C}$ \par
	\begin{algorithmic}[1]
		\FOR{$d_k \in \mathbb{D}$}
		\STATE initialize $\mathcal{G}=\{\theta_{\mathcal{F}},\theta_{\mathcal{E}},\theta_{\mathcal{D}}\} $ with $\mathcal{\widehat{G}}$
		\STATE $S^+_k=$ transform ($S$,$d_k$)\COMMENT{\textsl{As Fig. \ref{split}}}
		\STATE $\mathcal{G}=$  train ($\mathcal{G},S^+_k,T$)
		\STATE $p_k=$ validate ($\mathcal{G},S,T$)
		\COMMENT{\textsl{Val performance as reward}}
		\ENDFOR
		\STATE $\mathbb{P}=\{p_1,p_2...\}$
		\STATE $\mathcal{C}=$ train controller($\mathcal{C},\mathbb{D},\mathbb{P}$) 
		\COMMENT{\textsl{As Eq. \ref{equation:controller}}}
		\STATE $\mathbb{D}=$ update ($\mathcal{C},\mathbb{D}$)
		\COMMENT{\textsl{Same with NAO \cite{NAO}}}
		\STATE \textbf{return} $\mathbb{D},\mathbb{P}$
	\end{algorithmic}
\end{algorithm} 

%% file: illustration/datasets.tex
	
\newcommand{\tabincell}[2]{\begin{tabular}{@{}#1@{}}#2\end{tabular}}
\begin{table}[h]
	\centering
	\begin{tabular}{cccc}
		\toprule
		Dataset & Attribute  & \# Images   & \tabincell{c}{Cnt \\(Mean $\pm$ Std)} \\
		\toprule
		GCC & Syn  & 15,211 &501$\pm$718\\
		SHA & Real &482&501$\pm$456\\
		SHB & Real &716 &123$\pm$94\\
		QNRF & Real& 1,535	& 815$\pm$1176\\
		UCF-CC-50 & Real &50 &1,279$\pm$950\\
		WorldExpo'10 & Real &3,980&50$\pm$41\\
		\toprule
	\end{tabular}
	
	\caption{Statistics of five real-world (Real) datasets and one synthetic (Syn) dataset GCC used for experiments.}
	\label{table:datasets}
\end{table}

%% file: illustration/STOA.tex
\begin{table*}[h]
	\renewcommand*{\arraystretch}{1.15}
	\centering
	\begin{tabular}{cccccccccc}
		\toprule
		\multirow{2}{*}{Method}&\multicolumn{2}{c}{SHA}&\multicolumn{2}{c}{SHB}&\multicolumn{2}{c}{QNRF}
		&\multicolumn{2}{c}{UCF-CC-50} &WorldExpo'10 \\
		\cline{2-10} 
		~ &MAE& MSE & MAE& MSE& MAE& MSE & MAE& MSE & Avg. MAE\\
		\toprule
		NoAdpt \cite{cyclegan}& 160.0 &216.5& 22.8 & 30.6 & 275.5 & 458.5 & 487.2 & 689.0 & 42.8\\
		Cycle-GAN \cite{cyclegan} & 143.3 & 204.3 & 25.4 & 39.7 & 257.3 & 400.6 & 404.6 & 548.2 & 26.3\\ 
		SE Cycle-GAN \cite{GCC}& 123.4 & 193.4 & 19.9 & 28.3 & 230.4 &384.5 & 373.4 & 528.8 & 26.3\\ 
		SE Cycle-GAN(JT) \cite{cycleganjt} & 119.6 & 189.1 & 16.4 & 25.8 & 225.9 &385.7 & 370.2 & 512.0 & 24.4\\ 
		FSC \cite{FSC}& 129.3 & 187.6 & 16.9 & 24.7 & 221.2 & 390.2 & - & - & -\\ 
		FADA \cite{FADA}& - & - & 16.0 & 24.7 & - & - & - & - & 21.6\\ 
		GP \cite{GP}& 121.0 & 181.0 & 12.8 & 19.2 & 210.0 &351.0 & 355.0 & 505.0 & 20.4\\ 
		
		DACC \cite{gao2019domain}& 112.4 & 176.9 & 13.1 & 19.4 &211.7 & 357.9 & - & - &\textbf{17.4}\\ 
		\toprule
		BLA (ours) & \textbf{99.3} & \textbf{145.0} & \textbf{11.9} & \textbf{18.9} & \textbf{198.9} &\textbf{316.1}& \textbf{346.8} & \textbf{480.0} & 17.9\\ 
		\toprule
	\end{tabular}
	\caption{Comparison of our method with previous methods for synthetic-to-real adaptation. All methods employ VGG16 \cite{VGG} as backbone.}
	\label{table:STOA}
\end{table*}

%% file: illustration/vision_results.tex
\begin{figure*}[t]
	\begin{center}
		\includegraphics[width=0.93\textwidth]{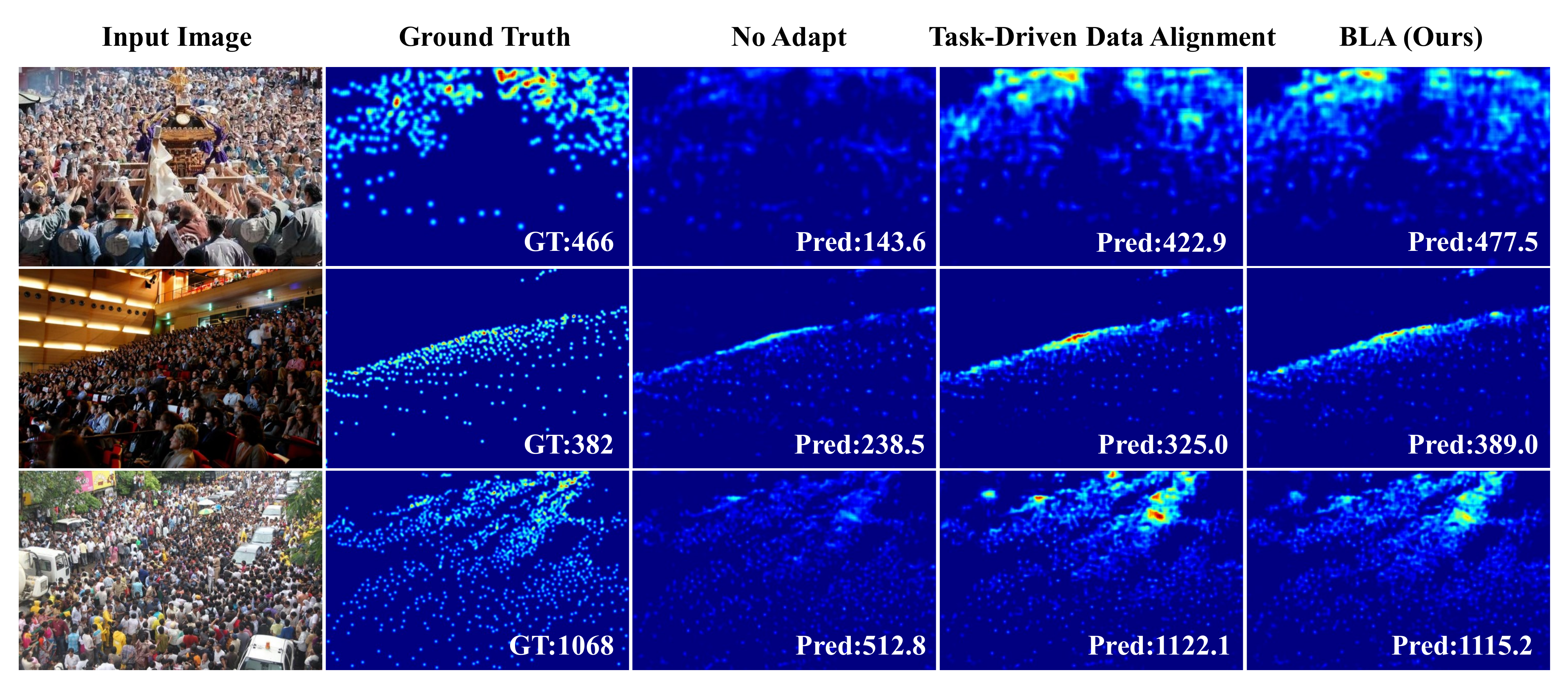}
	\end{center}
	   \vspace{-15pt}
	\caption{Qualitative adaptation results of images with different levels of density on SHA. 
	}
	\vspace{-10pt}
	\label{vision_results}
\end{figure*}

%% file: illustration/searched_d.tex

\begin{table}[h]
	\renewcommand*{\arraystretch}{1.1}
	\centering
	\begin{tabular}{cccc}
		\toprule
		\multirow{2}{*}{Dataset}&\multicolumn{3}{c}{Searched transform $d$} \\
		\cline{2-4} 
		~ &p$_\text{G}$& Scale factor & Angle\\
		\toprule
		SHA&0.78&0.77&$12^{\circ}$\\
		SHB&0.16&0.42&$3^{\circ}$\\
		QNRF&0.67&0.41&$25^{\circ}$\\
		UCF-CC-50&0.85&0.23&$4^{\circ}$\\
		WorldExpo'10&0.98&0.57&$17^{\circ}$\\
		\toprule
	\end{tabular}
	\caption{Searched transforms vary across different target datasets.}
	\label{table:searched_d}
	\vspace{-10pt}
\end{table}

%% file: illustration/different_dataset.tex
\begin{table}[h]
	\renewcommand*{\arraystretch}{1.15}
	\centering
	\begin{tabular}{ccccc}
		\toprule
		\multirow{2}{*}{Dataset}&\multicolumn{2}{c}{Searched $d$ for SHA}&\multicolumn{2}{c}{Searched $d$ for SHB}\\
		~&\multicolumn{2}{c}{[0.78, 0.77, $12^{\circ}$]}&\multicolumn{2}{c}{[0.16, 0.42, $3^{\circ}$]}\\
		\cline{2-5} 
		~ &MAE& MSE & MAE& MSE\\
		\toprule
		SHA& \textbf{99.3} &\textbf{145.0}& 104.3 & 153.7 \\ 
		SHB& 14.7 & 26.4 & \textbf{11.9} & \textbf{18.9} \\ 
		\toprule
	\end{tabular}
	\caption{Different datasets require different transformations for optimal results.}
	\vspace{-10pt}
	\label{table:different_dataset}
\end{table}

%% file: illustration/ablation.tex
\begin{table}[h]
	\centering
	\begin{tabular}{cccc}
		\toprule
		\tabincell{c}{Task-Driven \\  Data Alignment} &\tabincell{c}{Fine-Grained \\  Feature Alignment}   & MAE &MSE \\
		\toprule
		$\times$ & $\times$ & 134.7 & 210.9 \\
		\checkmark & $\times$ & 109.1 &153.8 \\
		$\times$ & \checkmark & 121.1&200.8 \\
		\checkmark & \checkmark& 99.3 & 145.0	\\
		\toprule
	\end{tabular}
	\caption{Effects of two levels of alignment. 
}
	\label{table:ablation}
\end{table}

%% file: illustration/drstdla.tex

\begin{table}[h]
	\centering
	\begin{tabular}{ccc}
		\toprule
		Method & MAE & MSE \\
		\toprule
		Style Transfer  & 119.4& 194.6 \\ 
		Domain Randomization  & 110.0 & 164.2 \\ 

		Task-Driven Data Alignment  & \textbf{99.3} & \textbf{145.0} \\
		\toprule
	\end{tabular}
	\caption{Effect of Task-Driven Data Alignment.}
	\label{table:drstdla}
\end{table}

%% file: illustration/adain.tex

\begin{figure}[h]
  \centering
  \includegraphics[scale=0.25]{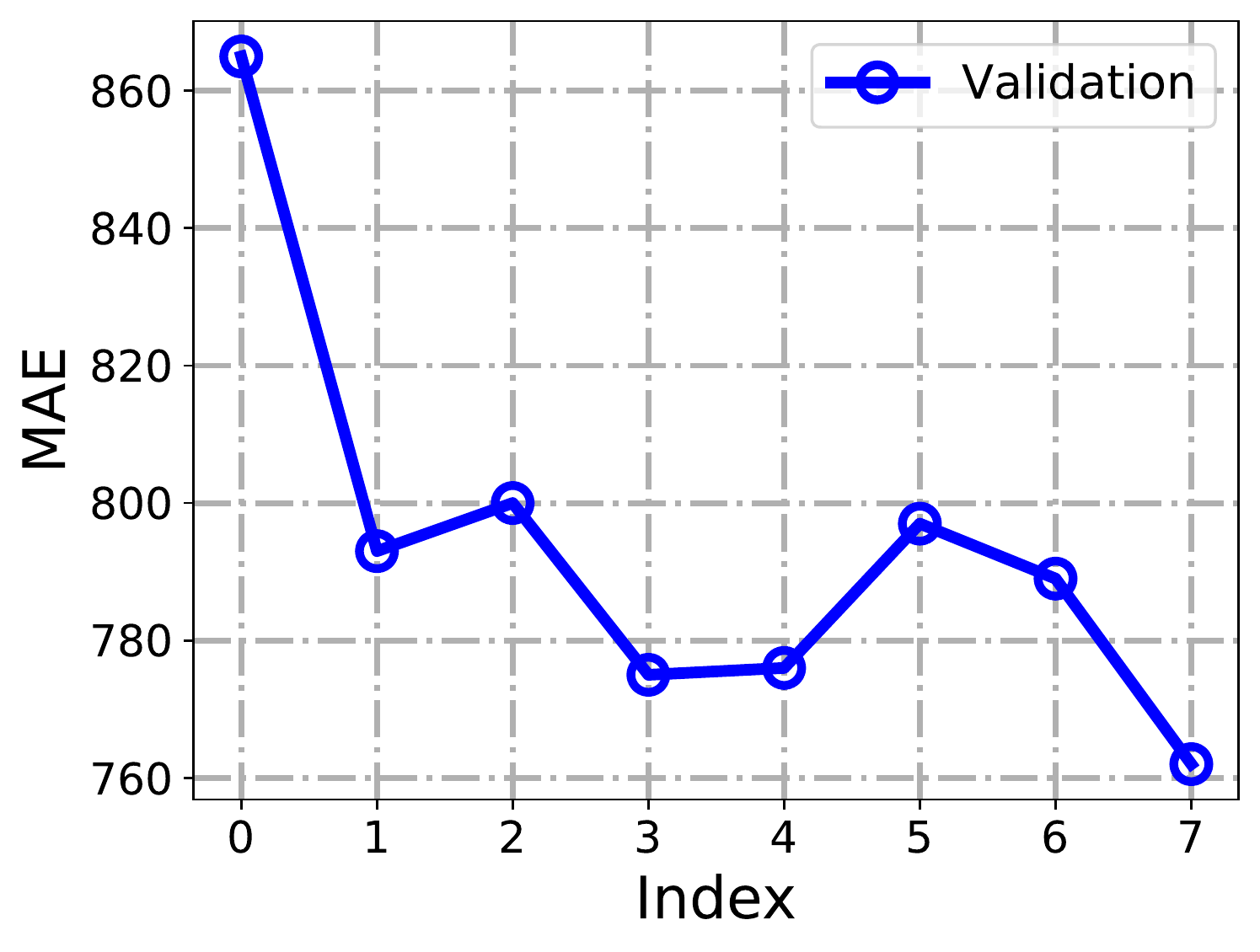}
  \hspace{1em}
  \includegraphics[scale=0.25]{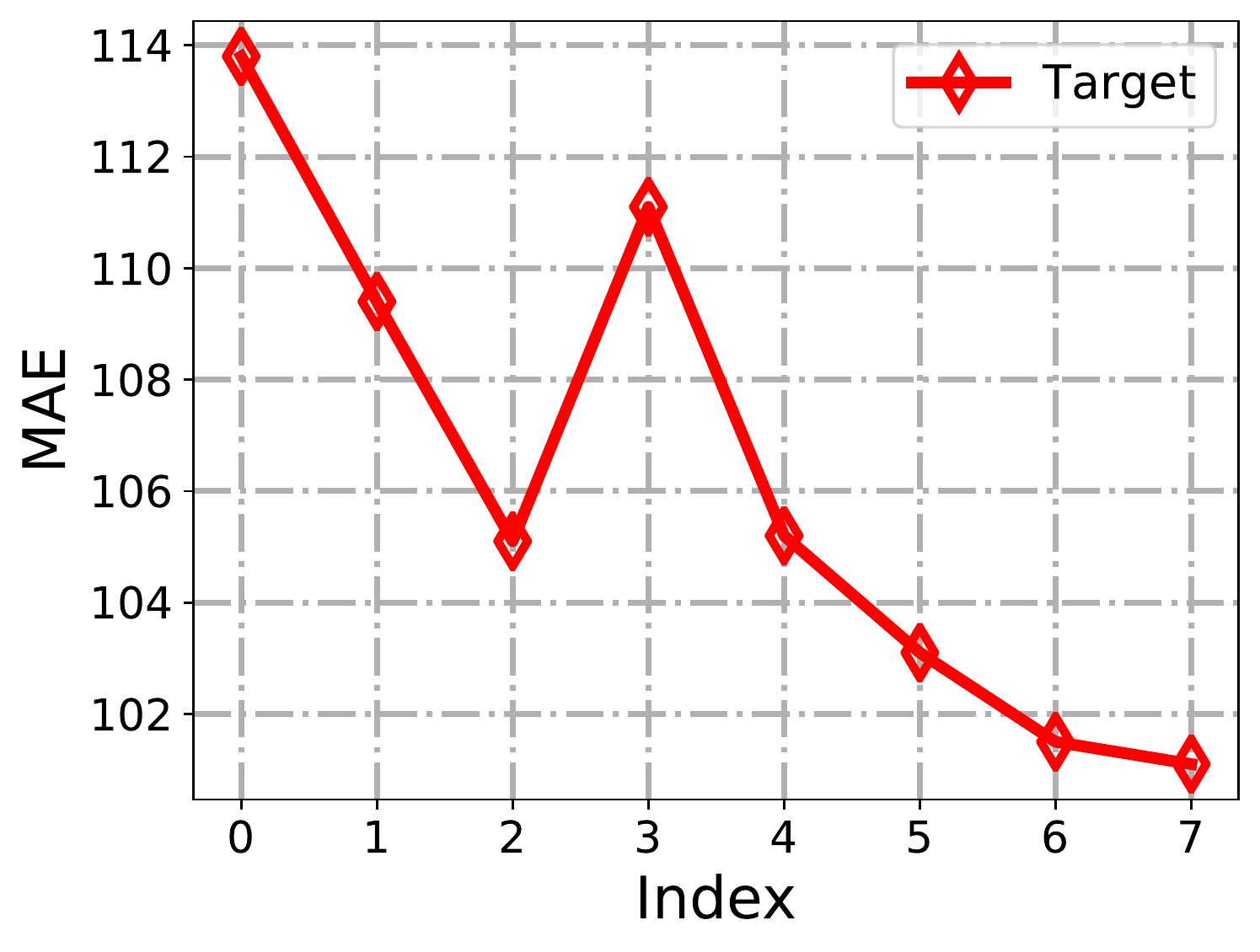}
  \caption{Comparison of validation performance on the generated set (left) and real target set (right) across different transforms w.r.t. MAE. The similar trend verifies the effect of our validation feature generator. 
  }
\vspace{-10pt}
  \label{adain}
\end{figure}

%% file: illustration/grid.tex

\begin{table}[h]
	\centering
	\begin{tabular}{ccc}
		\toprule
		Grid size $G$ & MAE & MSE \\
		\toprule
		(2,2) & 138.1 & 222.9 \\
		(4,4) & 130.5 & 205.4 \\
		(8,8) & 129.0 & 206.3 \\
		(16,16) & 121.1 & 200.8 \\
		(32,32) & 141.4 & 223.6 \\
		\toprule
	\end{tabular}
	\caption{Impact of local grid size $G$ used in fine-grained feature alignment.
}
	\label{table:grid}
\end{table}
